\documentclass{article}
\usepackage{spconf,amsmath,graphicx,tikz,pgfplots,bbding,multirow,subcaption,amsfonts,float,booktabs,hyperref}
\usepackage[nobiblatex]{xurl} 
\usepackage{xpatch} 

\definecolor{lightblue}{HTML}{a3b4d8}
\definecolor{charcoal}{HTML}{264653}
\definecolor{persiangreen}{HTML}{2A9D8F}
\definecolor{orangeyellow}{HTML}{E9C46A}
\definecolor{sandybrown}{HTML}{F4A261}
\definecolor{burntsienna}{HTML}{E76F51}

\newcommand{\pastref}{\hat{\mathbf{x}}_{p}}      
\newcommand{\futureref}{\hat{\mathbf{x}}_{f}}    
\newcommand{\pastflow}{\mathbf{v}_{p}}           
\newcommand{\futureflow}{\mathbf{v}_{f}}         
\newcommand{\pred}{\tilde{\mathbf{x}}_t}         
\newcommand{\bbeta}{\boldsymbol{\beta}}          
\newcommand{\balph}{\boldsymbol{\alpha}}         

\title{AIVC: Artificial Intelligence based Video Codec}
\name{Théo Ladune and Pierrick Philippe}
\address{\texttt{theo.ladune@orange.com}, Orange, France}

\begin{document}
%
\maketitle
\begin{abstract}
This paper introduces AIVC, an end-to-end neural video codec. It is based on two
conditional autoencoders MNet and CNet, for motion compensation and coding. AIVC
learns to compress videos using any coding configurations through a single
end-to-end rate-distortion optimization. Furthermore, it offers performance
competitive with the recent video coder HEVC under several established test
conditions. A comprehensive ablation study is performed to evaluate the benefits
of the different modules composing AIVC. The implementation is made available at
\small{\url{https://orange-opensource.github.io/AIVC/}}.
\end{abstract}
%

\section{Introduction \& related works}

Digital technologies are becoming an ever-growing part of our daily life. This
has an important environmental impact, caused by a rising number of devices
(data centers, networking equipment, user terminals). In particular, video
streaming causes a significant share of this impact as it represents more than
75~\% of overall Internet traffic \cite{cisco}. Reducing the size of the videos
exchanged over the Internet thus alleviates some inconveniences of digital
technologies.

Standardization organisms such as MPEG and ITU have released several video
coding standards (AVC \cite{avc} in 2003, HEVC \cite{hevc} in 2013 and VVC
\cite{vvc} in 2020), reducing the size of videos while maintaining an acceptable
visual quality. Recently, neural-based coders have been studied by the
compression community. In the span of a few years, they have reached image
coding performance on par with VVC \cite{cheng2020learned}. Yet, video coding
remains a challenging task for neural coders, due to the additional temporal
dimension.
\newline

Conventional and neural video coders rely on similar techniques to remove
temporal redundancies in a video. First, a temporal prediction is computed at
the decoder. Then, only the unpredicted part is sent from the encoder to the
decoder. Many previous works have refined these two steps. For instance,
temporal prediction is performed either in the spatial
\cite{Agustsson_2020_CVPR} or feature \cite{guo2021learning} domain. Similarly,
the unpredicted part is computed in the spatial \cite{9506275} or feature
\cite{ladune2021conditional} domain.

Yet, these refinements are often evaluated under particular test conditions
which are arguably different from the requirements of the industry. Most
previous works focus on the low-delay P configuration
\cite{Agustsson_2020_CVPR,guo2021learning} (used for videoconferencing) and omit
the Random Access configuration (used for streaming at large). Furthermore, most
neural codecs \cite{hu2020improving,hu2021fvc} are assessed using an I frame
period shorter (\textit{e.g.} 10 or 12 frames, regardless of the video
framerate) than expected by the Common Test Conditions of modern video coders
such as those defined for HEVC or VVC \cite{HEVC_CTC}. Consequently, neural
coder performance is not accurately evaluated.
\newline

This paper introduces AIVC, an AI-based Video Codec featuring both
conditional coding \cite{ladune2021conditional} and Skip \cite{9287049}
mechanisms. AIVC is designed to be a versatile codec, able to implement any desired
coding configuration. It is evaluated under test conditions which strive to
reconcile the learned and conventional video coding community. On the one hand, the
CLIC 2021 \cite{CLIC21} test sequences and quality metric (MS-SSIM) are used. On
the other hand, the HEVC Test Model (HM) serves as anchor and 3 configurations
are evaluated: Random Access (one I frame per second), Low-delay P and All
Intra. Our contributions are summarized as follows:
\begin{enumerate}
    \itemsep0em
    \item We propose an easy-to-train architecture, competitive with HEVC under
    various test conditions;
    \item We provide comprehensive experimental results justifying all components composing AIVC;
    \item We publicly release the trained models \cite{AIVCwebsite}.
\end{enumerate}

\section{Proposed system}

\begin{figure*}[ht]
    \centering
    \includegraphics[width=0.97\textwidth]{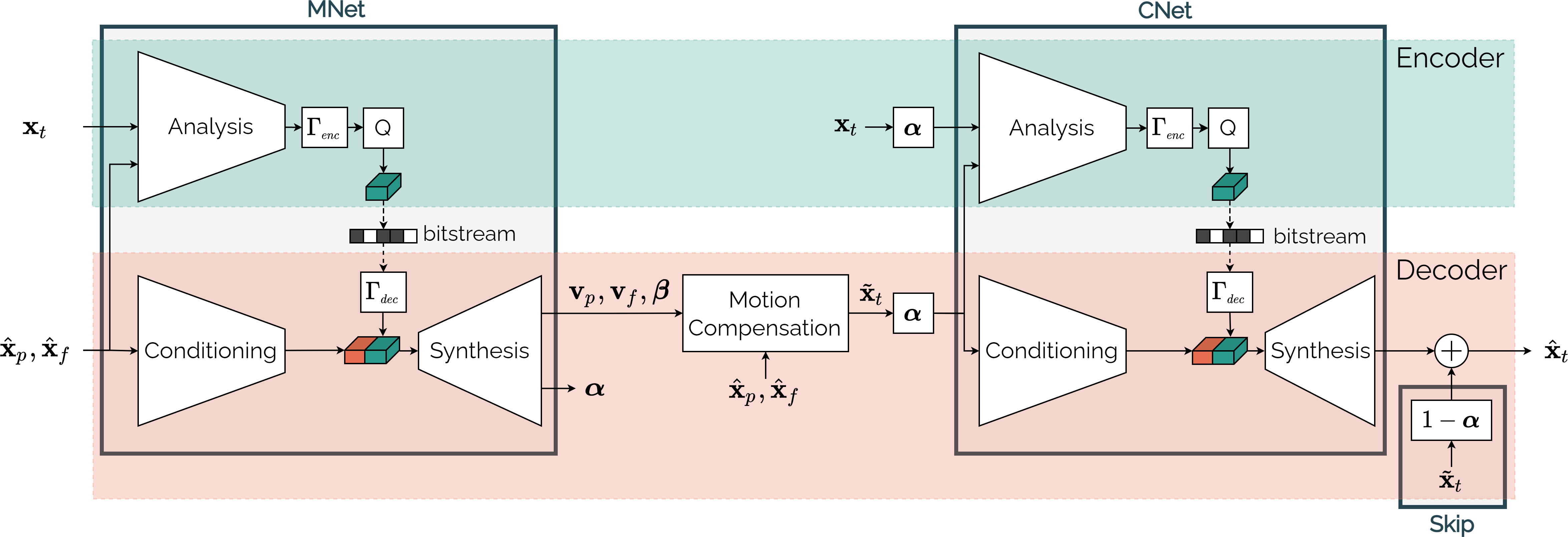}
    \caption{Block diagram of the coding scheme.}
    \label{fig:global_diagram}
\end{figure*}

\subsection{System overview}

Let us consider a video as a sequence of frames, where each frame $\mathbf{x}_t$
is a $3 \times H \times W$ tensor\footnote{A bilinear upsampling is used to
convert YUV 420 data into YUV 444.}. Similarly to conventional video codecs,
AIVC processes a frame while using information from up to 2 already transmitted
frames, called reference frames. These two references (one past and one future)
are denoted $\pastref$ and $\futureref$. If the coding of $\mathbf{x}_t$
exploits both references, $\mathbf{x}_t$ is called a B frame. A P frame uses a
single reference ($\futureref = 0$), while an I frame uses no references
($\pastref = \futureref = 0$).

AIVC processes a frame $\mathbf{x}_t$ following the coding pipeline shown in
Fig. \ref{fig:global_diagram}. First, motion information is computed and sent by
a neural network MNet. This information comprises two pixel-wise motion fields
$\pastflow, \futureflow$ (for the past and future references) and one pixel-wise
prediction weighting $\bbeta$. Then, a bi-directional motion compensation
algorithm computes a temporal prediction $\pred$:
\begin{equation}
    \pred = \bbeta w\left(\pastref,\pastflow\right) + \left(1 - \bbeta\right) w\left(\futureref,\futureflow\right),
\end{equation}
with $w$ a bilinear warping. Finally, the unpredicted part of $\mathbf{x}_t$ is
sent using a second neural network CNet. That is, the coding of
$\mathbf{x}_t$ is performed conditionally to its prediction $\pred$.

While some previous works \cite{Agustsson_2020_CVPR} require a separate I frame
network, AIVC compresses all frame types (I, P \& B) identically, simply zeroing
the unavailable references.

\subsection{Content adaptation with Skip mode}

Conventional video coders (\textit{e.g.} HEVC, VVC) are characterized by the
important number of available coding modes \textit{i.e.} different ways of
processing a set of pixels. This allows performing operations adapted to
different video content and leads to compelling performance. Following
this idea, AIVC features an additional coding mode called Skip mode
\cite{9287049}.

Skip mode offers the possibility to shortcut CNet, by directly using areas of
the prediction $\pred$ as the decoded frame. This coding mode is particularly
convenient for well-predicted areas. The choice between CNet and Skip is
arbitrated pixel-wise through a multiplication by $\balph$, the mode selection
(see Fig. \ref{fig:global_diagram}). As $\balph$ must be known at the decoder,
it is computed and conveyed by MNet, alongside the motion information. Adding
Skip mode improves the performance of AIVC as it can better adapt to the video
to be compressed. Furthermore, Skip mode eases the training convergence since it
fosters the learning of a relevant prediction and accurate motion information.

\subsection{Conditional coding for MNet \& CNet}

\begin{figure*}
    \centering
    \begin{subfigure}[t]{0.32\linewidth}
        \centering
        \includegraphics[width=0.987\linewidth]{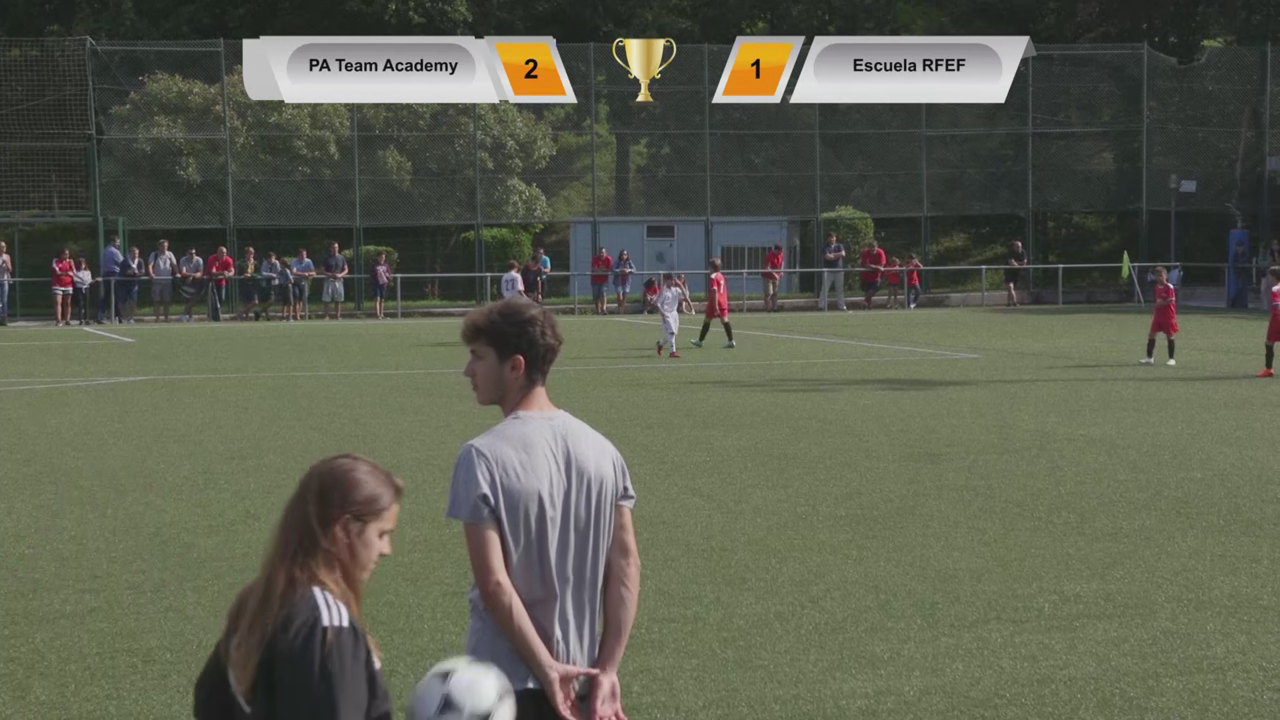}
        \caption{Input frame $\mathbf{x}_t$}
        \label{fig:visual_example:x}
    \end{subfigure}
    \begin{subfigure}[t]{0.32\linewidth}
        \centering
        \includegraphics[width=0.987\linewidth]{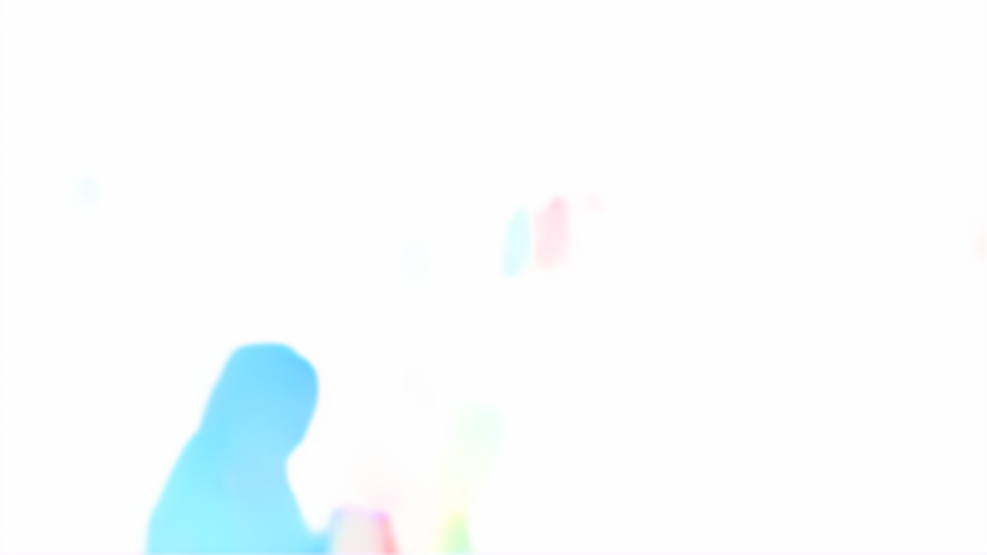}
        \caption{Motion field $\mathbf{v}_p$}
        \label{fig:visual_example:past_flow}
    \end{subfigure}
    \begin{subfigure}[t]{0.32\linewidth}
        \centering
        \includegraphics[width=0.987\linewidth]{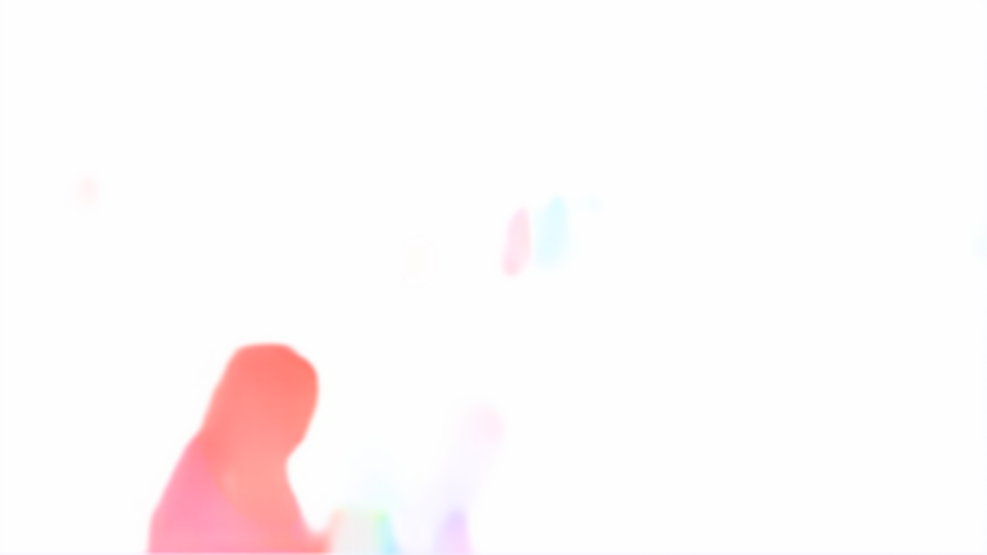}
        \caption{Motion field $\mathbf{v}_f$}
        \label{fig:visual_example:future_flow}
    \end{subfigure}

    \begin{subfigure}[t]{0.32\linewidth}
        \centering
        \includegraphics[width=0.987\linewidth]{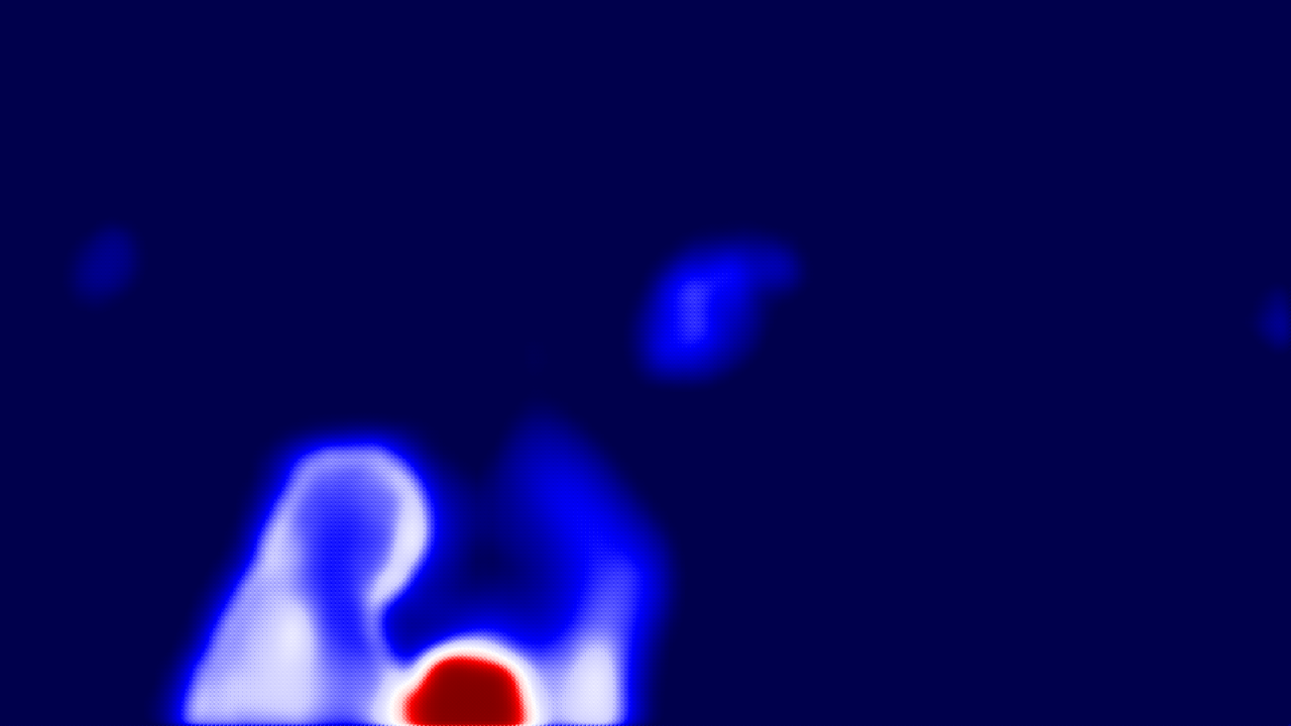}
        \caption{Coding mode $\balph$. Blue: Skip, red: CNet.}
        \label{fig:visual_example:alpha}
    \end{subfigure}
    \begin{subfigure}[t]{0.32\linewidth}
        \centering
        \includegraphics[width=0.987\linewidth]{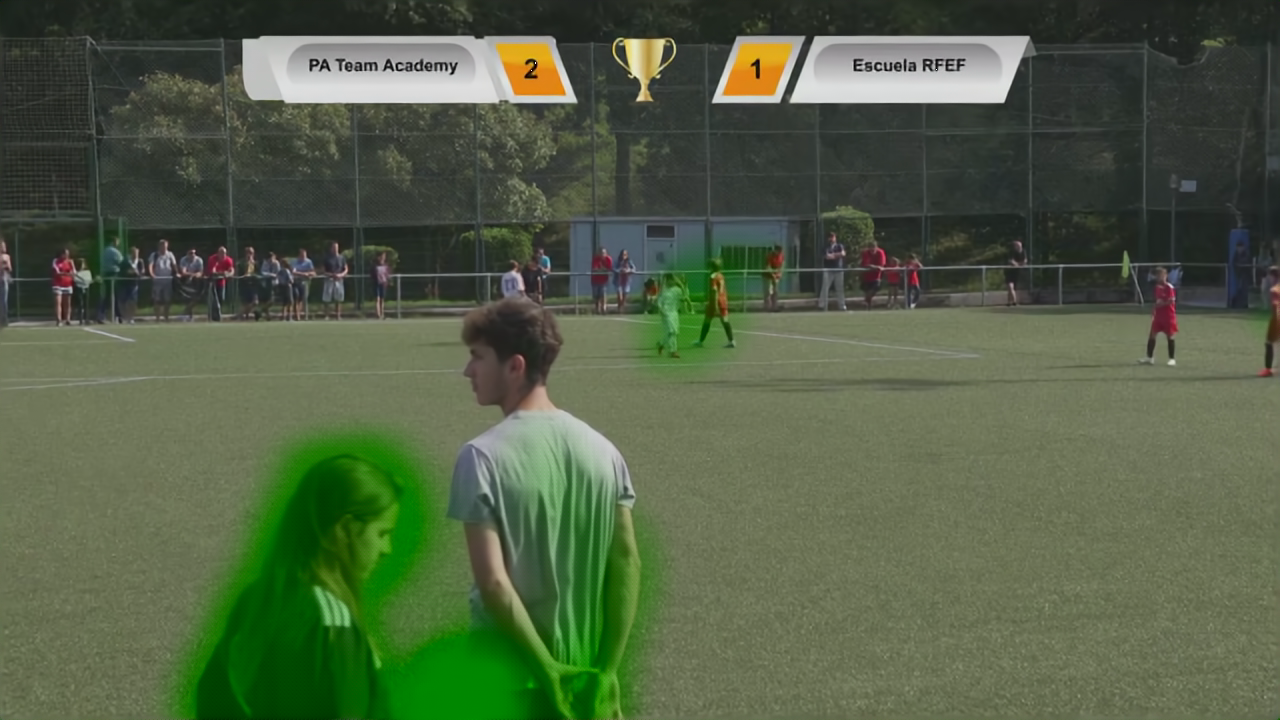}
        \caption{Skip contribution $\left(1 - \balph\right) \odot \tilde{\mathbf{x}}_t$}
        \label{fig:visual_example:skip_contrib}
    \end{subfigure}
    \begin{subfigure}[t]{0.32\linewidth}
        \centering
        \includegraphics[width=0.987\linewidth]{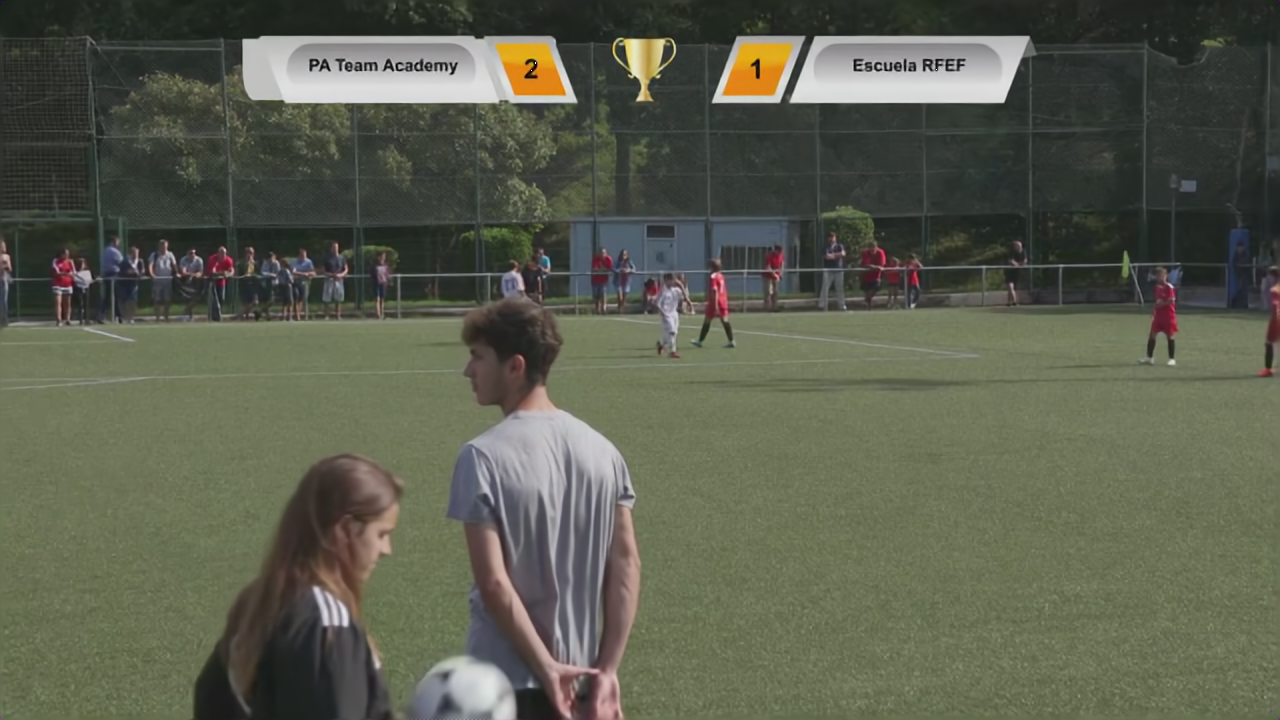}
        \caption{Decoded frame $\hat{\mathbf{x}}_t$}
        \label{fig:visual_example:x_hat}
    \end{subfigure}
    \caption{Visual examples on a B frame from the CLIC 2021 sequence
    \textit{Sports\_1080P-6710}.}
    \label{fig:visual_example}
\end{figure*}

Despite different roles, CNet and MNet share the same architecture called
conditional coding (CC) \cite{ladune2021conditional,thesistheo}, to exploit
decoder-side information as much as possible. To this effect, CC adds a
\textit{conditioning} transform to the usual analysis-synthesis convolutional
autoencoder \cite{balle2018variational}. At the decoder, the conditioning
transform computes a conditioning latent variable representing the available
decoder-side information. At the encoder, the analysis transform identifies the
information missing at the decoder. It is fed with the encoder-side and
decoder-side data to compute an analysis latent variable, which is then sent to
the decoder. Consequently, the encoder (subject to a rate constraint) transmits
only the unpredicted part of $\mathbf{x}_t$. Finally the synthesis transform
processes both latent variables to obtain the desired output.

CC is used as a generic architecture to exploit decoder-side information
regardless of the information nature. For instance, MNet leverages CC by using
the image-domain data (reference frames) to retrieve information about the
motion information and the coding mode selection. Compared to residual coding,
CC offers a richer non-linear mixture in the latent domain which results in
better compression performance.

\subsection{Variable quantization gains}

The importance of the analysis latent variable (both for MNet and CNet) depends
on the availability of the reference frames. When no reference is available, the
conditioning transform cannot extract relevant decoder-side information.
Consequently, all the required information is conveyed through the analysis
transform. To better adapt to the importance of the analysis latent variable,
AIVC features different quantization gains based on the frame type.

Quantization gains are derived from the multi-rate codec proposed in
\cite{Cui_2021_CVPR}. For each frame type $f \in \left\{I, P, B\right\}$, a
feature-wise pair of gains $(\Gamma_{f}^{enc},\Gamma_f^{dec})$ is learned. Each
gain $\Gamma \in \mathbb{R}^{F}$, with $F$ the number of channels of the
analysis latent variable. Gains multiply the analysis latent variable before and
after an unitary quantizer.

\subsection{Architecture details}

MNet and CNet analysis and synthesis transforms are implemented using the
convolutional autoencoder architecture proposed in \cite{cheng2020learned}. They
feature attention modules, residual blocks and the hyperprior mechanism. The
conditioning transform of MNet and CNet replicates the architecture of the
analysis transform. As a result, AIVC has 50 million parameters.

\section{Experimental results}

\subsection{Training}

\begin{figure}[H]
    \centering
    \begin{subfigure}[b]{0.45\linewidth}
        \centering
        \includegraphics[width=\linewidth]{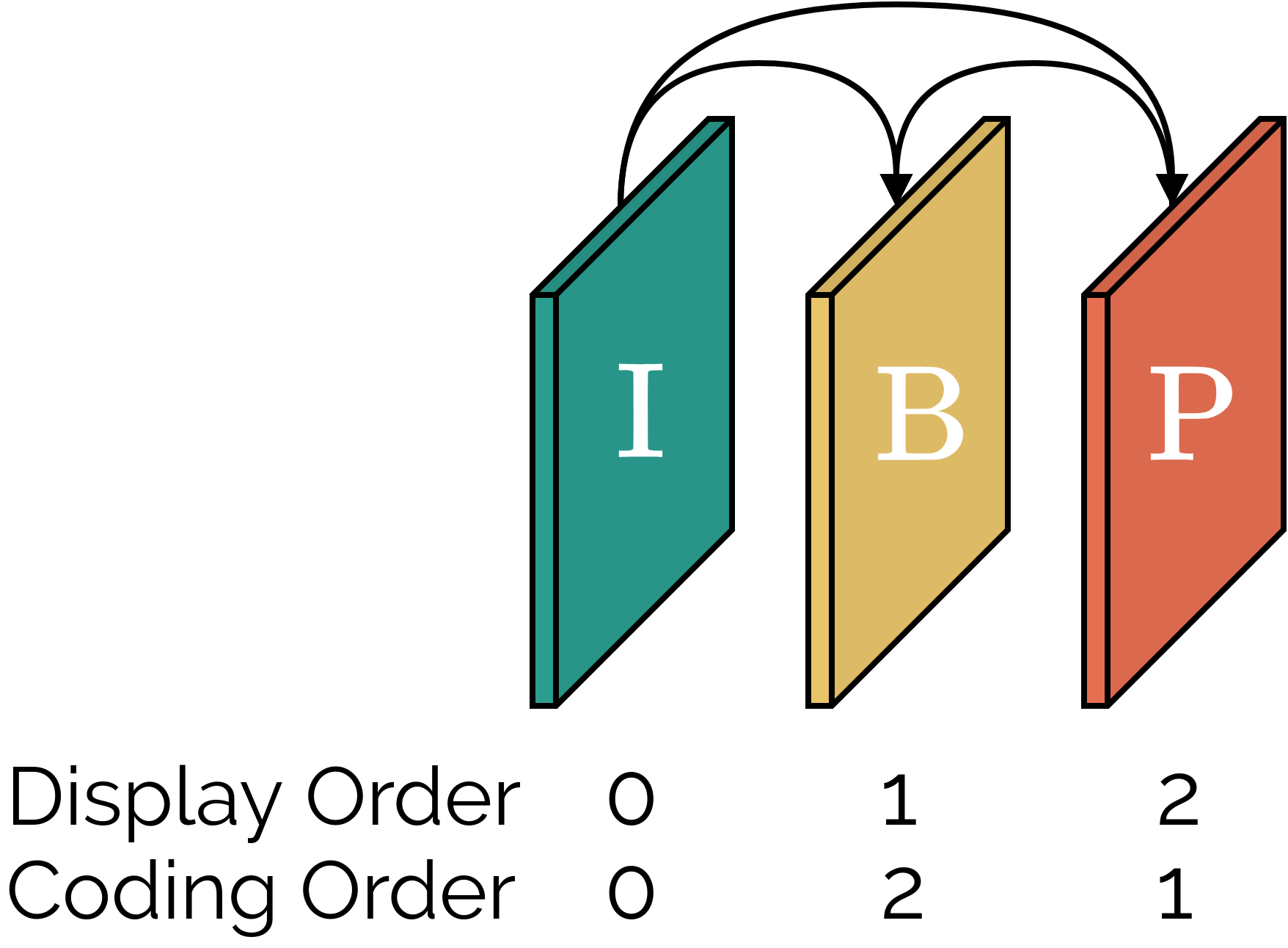}
        \caption{Training configuration.}
        \label{fig:training_config}
    \end{subfigure}\quad
    \begin{subfigure}[b]{0.5\linewidth}
        \includegraphics[width=\linewidth]{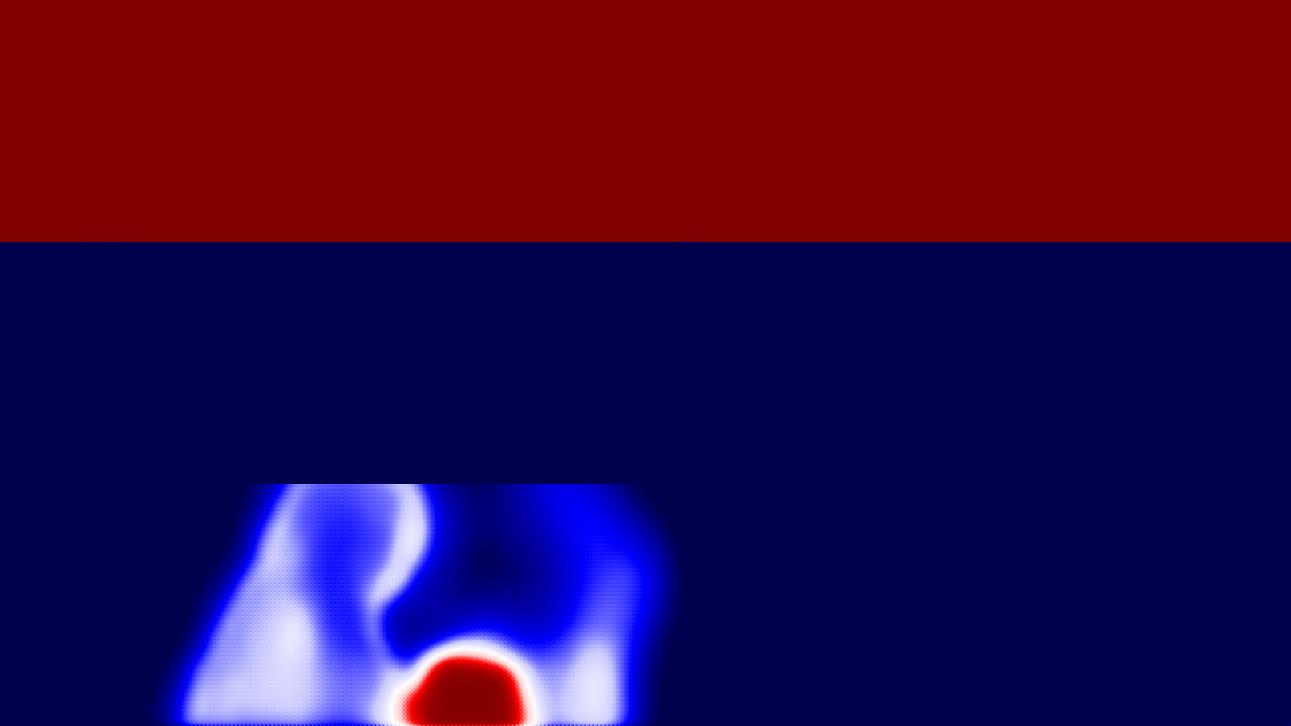}
        \caption{$\balph$ during the first iterations. Blue: Skip, red: CNet.}
        \label{fig:training_alpha}
    \end{subfigure}
    \caption{Additional details on the training stage.}
\end{figure}

AIVC is designed to code any configuration composed of I, P and B frames. As
such, a configuration featuring all 3 frame types is used for training (Fig.
\ref{fig:training_config}). For each training iteration, the 3 frames are coded
and gradient descent is used to minimize the loss function:
\begin{equation}
    \mathcal{L}_\lambda = \sum_t \mathrm{D}\left(\mathbf{x}_t, \hat{\mathbf{x}}_t\right)
    + \lambda \left(R_m + R_c\right).
\end{equation}

The distortion $\mathrm{D}$ is based on MS-SSIM to comply with the CLIC 2021
test conditions \cite{CLIC21}. The rate constraint $\lambda$ balances
$\mathrm{D}$ with MNet rate $R_m$ and CNet rate $R_c$. During training, the
entropy of the analysis latent variables acts as a proxy for the rate
\cite{balle2018variational}. Different $\lambda$ are used to obtain systems with
different rate targets. Training examples are extracted from several datasets:~KoNVid\_1k
\cite{hosu2017konstanz}, YouTube-NT \cite{DBLP:conf/iclr/YangYMM21}, YUV\_4K
\cite{DBLP:conf/mmsys/MercatVV20} and CLIC \cite{CLIC21}.

Training AIVC does not require auxiliary losses
\cite{DBLP:conf/accv/GolinskiPYSC20} or pre-trained motion components
\cite{lu:dvc}. Instead, Skip mode fosters the learning of relevant motion
information ($\pastflow, \futureflow$ and $\bbeta$) and coding mode selection
$\balph$. However, during the first training iterations CNet is not yet ready to
compete with Skip. As such, $\balph$ is forced to zero (Skip) and one (CNet) on
some areas of the frame (Fig. \ref{fig:training_alpha}).

\subsection{Visual examples}

\definecolor{lavenderpurple}{rgb}{0.59, 0.48, 0.71}
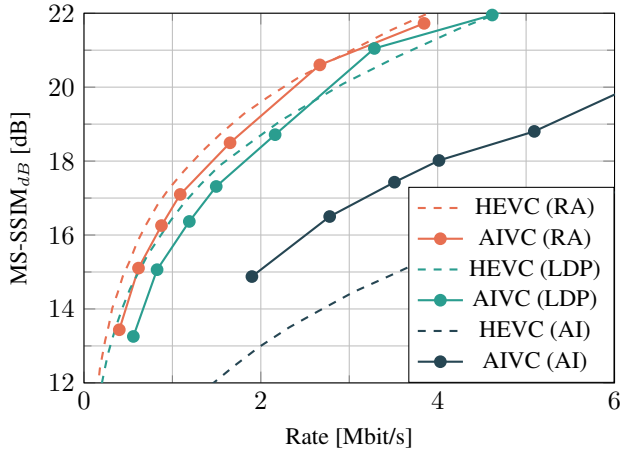
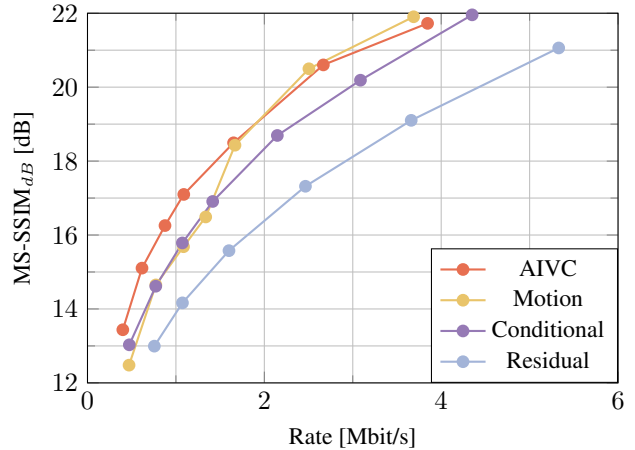
\begin{figure*}
    \centering
    \begin{subfigure}{0.49\textwidth}
        \centering
        \begin{tikzpicture}
            \begin{axis}[
                grid= both,
                xlabel = {\small Rate [Mbit/s]},
                ylabel = {\small $\text{MS-SSIM}_{dB}$ [dB]} ,
                xmin = 0, xmax = 6,
                ymin = 12, ymax = 22,
                ylabel near ticks,
                xlabel near ticks,
                height=6.5cm,
                width=0.99\linewidth,
                xtick distance={2},
                ytick distance={2},
                minor y tick num=1,
                minor x tick num=1,
                legend style={
                    at={(1,0)},
                    anchor=south east,
                }
            ]

                \addplot[dashed, thick, burntsienna,] table {data/clic21/hm_ra.txt};
                \addlegendentry{\small HEVC (RA)}

                \addplot[solid, thick, burntsienna, mark=*, mark options={solid}] table {data/clic21/aivc_ra.txt};
                \addlegendentry{\small AIVC (RA)}

                \addplot[dashed, thick, persiangreen] table {data/clic21/hm_ldp.txt};
                \addlegendentry{\small HEVC (LDP)}

                \addplot[solid, thick, persiangreen, mark=*, mark options={solid}] table {data/clic21/aivc_ldp.txt};
                \addlegendentry{\small AIVC (LDP)}

                \addplot[dashed, thick, charcoal,] table {data/clic21/hm_ai.txt};
                \addlegendentry{\small HEVC (AI)}

                \addplot[solid, thick, charcoal, mark=*, mark options={solid}] table {data/clic21/aivc_ai.txt};
                \addlegendentry{\small AIVC (AI)}
            \end{axis}
        \end{tikzpicture}
        \caption{Compression performance of AIVC against HEVC.}
        \label{fig:rd_results}
    \end{subfigure}\quad
    \begin{subfigure}{0.49\textwidth}
        \centering
        \begin{tikzpicture}
            \begin{axis}[
                grid= both,
                xlabel = {\small Rate [Mbit/s]},
                ylabel = {\small $\text{MS-SSIM}_{dB}$ [dB]} ,
                xmin = 0, xmax = 6,
                ymin = 12, ymax = 22,
                ylabel near ticks,
                xlabel near ticks,
                height=6.5cm,
                width=0.99\linewidth,
                xtick distance={2},
                ytick distance={2},
                minor y tick num=1,
                minor x tick num=1,
                legend style={
                    at={(1,0)},
                    anchor=south east,
                }
            ]
                \addplot[thick, burntsienna, mark=*, mark options={solid}] table {data/clic21/aivc_ra.txt};
                \addlegendentry{\small AIVC}

                \addplot[thick, orangeyellow, mark=*, mark options={solid}] table {data/ablation/cc_gain_motion.txt};
                \addlegendentry{\small Motion}

                \addplot[thick, lavenderpurple, mark=*, mark options={solid}] table {data/ablation/cc_gain.txt};
                \addlegendentry{\small Conditional}

                \addplot[thick, lightblue, mark=*, mark options={solid}] table {data/ablation/res.txt};
                \addlegendentry{\small Residual}

            \end{axis}
        \end{tikzpicture}
        \caption{Ablation study (Random Access configuration).}
        \label{fig:ablation:curve}
    \end{subfigure}
    \caption{Rate-distortion curves on the CLIC 2021 validation set.  $\text{MS-SSIM}_{dB} = -10 \log_{10} \left(1 - \text{MS-SSIM}\right)$}
\end{figure*}
Figure \ref{fig:visual_example} illustrates the processing of a B frame
$\mathbf{x}_t$ (Fig. \ref{fig:visual_example:x}). First, MNet computes and
transmits two motion fields (Fig. \ref{fig:visual_example:past_flow} and
\ref{fig:visual_example:future_flow}) allowing a temporal prediction
$\pred$ to be computed, leveraged by two coding modes:~Skip and CNet. These modes are
arbitrated by $\balph$ (Fig. \ref{fig:visual_example:alpha}). Skip mode (Fig.
\ref{fig:visual_example:skip_contrib}) is a direct copy of $\pred$ which is
mostly used for the well-predicted areas \textit{e.g.} slow moving objects.
Areas which do not rely on Skip are conveyed by CNet. Finally, both coding mode
contributions are added to obtain the decoded frame (Fig.
\ref{fig:visual_example:x_hat}).

Although the optimization process is driven only by the rate-distortion
objective, MNet learns relevant motion fields and coding mode selection. They
present accurate values and edges as well as smooth low-frequency areas suited
for low-rate transmission. Supplementary animated illustrations are provided
alongside the models \cite{AIVCwebsite}.

\subsection{Rate-distortion performance}

AIVC performance is evaluated against the HEVC test Model (HM) 16.22 under 3
configurations:~Random Access (RA) which features I, P and B frames, Low-delay P
(LDP) with one initial I frame followed by P frames and All Intra (AI)
consisting only of I frames. Test sequences are from the CLIC 2021 validation
set and the quality metric is MS-SSIM.


Figure \ref{fig:rd_results} presents the results. AIVC is competitive with HEVC
for RA and LDP at higher rates, while it is slightly worse at lower rates.
Moreover, AIVC significantly outperforms HEVC for AI coding. These results
validate the design choices made for AIVC. Yet, further work should focus on
enhancing the motion-related components of AIVC to outperform HEVC RA results.

\subsection{Ablation results}

This section illustrates the benefits brought by different
components of AIVC. Figure \ref{fig:ablation:curve} shows the rate-distortion
performance of the different configurations presented in Table
\ref{fig:ablation:table}.
\newline
\textbf{Residual} is the most basic configuration. It does not feature motion
compensation nor Skip ($\pastflow = \futureflow = \balph = 0$) and the
prediction $\pred$ is the average of the reference frames. CNet is implemented
as a normal autoencoder (\textit{i.e.} no conditioning transform) conveying the
prediction error $\mathbf{x}_t - \pred$.
\newline
\textbf{Conditional} simply modifies CNet, replacing residual coding by
conditional coding and adding frame type adapted quantization gains (no motion
compensation is present). The significant increase in performance highlights the
benefits of replacing residual coding with conditional coding.
\newline
\textbf{Motion} adds MNet and motion compensation yielding a more accurate
prediction. MNet is a normal autoencoder (\textit{i.e.} no conditioning
transform) and Skip mode is not used. The introduction of motion information
improves high-rate results but does not enhance performance at lower rates.
\newline
\textbf{AIVC} shows the relevance of Skip mode and conditional
coding for MNet which yields better results for lower rates.

\begin{table}[hb]
    \centering
    \caption{Ablation configurations. CC stands for conditional coder and AE for autoencoder.}
    \begin{tabular}{l|c c c c}
        Name  & CNet & MNet   & Motion comp.& Skip \\
        \midrule
        \midrule
        AIVC & CC & CC & \Checkmark & \Checkmark \\
        Motion & CC & AE & \Checkmark & \\
        Conditional & CC & & & \\
        Residual & Residual &  & & \\
    \end{tabular}
    \label{fig:ablation:table}
\end{table}

\section{Conclusion}

This paper presents AIVC, a learned video coder able to compress videos using
any coding configuration composed of I, P and B frames. AIVC is shown to be
competitive with the best implementation of HEVC under several test conditions.
Finally an ablation study highlights the benefits of each component \textit{e.g.} conditional
coding and Skip mode.

Although AIVC offers compelling performance, it remains challenging for neural
codecs to outperform modern conventional codecs (HEVC and VVC) especially at
lower rates. We believe that the introduction of additional coding modes
(similar to Skip mode) would improve neural codec results. Moreover the
experimental results provided highlight the relative weakness of the motion
component, which needs to be refined to obtain better performance.


\newpage

\bibliographystyle{IEEEbib}
\bibliography{refs}

\begin{thebibliography}{10}

\bibitem{cisco}
``Cisco predicts more ip traffic in the next five years than in the history of
  the internet,'' 2017.

\bibitem{avc}
T.~Wiegand, G.J. Sullivan, G.~Bjontegaard, and A.~Luthra,
\newblock ``Overview of the h.264/avc video coding standard,''
\newblock {\em IEEE Transactions on Circuits and Systems for Video Technology},
  2003.

\bibitem{hevc}
Gary~J. Sullivan, Jens-Rainer Ohm, Woo-Jin Han, and Thomas Wiegand,
\newblock ``Overview of the high efficiency video coding (hevc) standard,''
\newblock {\em IEEE Transactions on Circuits and Systems for Video Technology},
  2012.

\bibitem{vvc}
Benjamin Bross, Jianle Chen, Jens-Rainer Ohm, Gary~J. Sullivan, and Ye-Kui
  Wang,
\newblock ``Developments in international video coding standardization after
  avc, with an overview of versatile video coding (vvc),''
\newblock {\em Proceedings of the IEEE}, 2021.

\bibitem{cheng2020learned}
Zhengxue Cheng, Heming Sun, Masaru Takeuchi, and Jiro Katto,
\newblock ``Learned image compression with discretized gaussian mixture
  likelihoods and attention modules,''
\newblock in {\em 2020 {IEEE/CVF} Conference on Computer Vision and Pattern
  Recognition}.

\bibitem{Agustsson_2020_CVPR}
Eirikur Agustsson, David Minnen, Nick Johnston, Johannes Balle, Sung~Jin Hwang,
  and George Toderici,
\newblock ``Scale-space flow for end-to-end optimized video compression,''
\newblock in {\em Proceedings of the IEEE/CVF Conference on Computer Vision and
  Pattern Recognition}, 2020.

\bibitem{guo2021learning}
Zongyu Guo, Runsen Feng, Zhizheng Zhang, Xin Jin, and Zhibo Chen,
\newblock ``Learning cross-scale prediction for efficient neural video
  compression,'' 2021.

\bibitem{9506275}
David Alexandre, Hsueh-Ming Hang, Wen-Hsiao Peng, and Marek Domański,
\newblock ``Deep video compression for interframe coding,''
\newblock in {\em 2021 IEEE International Conference on Image Processing
  (ICIP)}.

\bibitem{ladune2021conditional}
Th{{\'{e}}}o Ladune, Pierrick Philippe, Wassim Hamidouche, Lu~Zhang, and
  Olivier D{{\'{e}}}forges,
\newblock ``Conditional coding for flexible learned video compression,''
\newblock in {\em Neural Compression: From Information Theory to Applications
  -- Workshop @ ICLR 2021}.

\bibitem{hu2020improving}
Zhihao Hu, Zhenghao Chen, Dong Xu, Guo Lu, Wanli Ouyang, and Shuhang Gu,
\newblock ``Improving deep video compression by resolution-adaptive flow
  coding,''
\newblock in {\em Computer Vision - {ECCV} 2020 - 16th European Conference}.

\bibitem{hu2021fvc}
Zhihao Hu, Guo Lu, and Dong Xu,
\newblock ``{FVC:} {A} new framework towards deep video compression in feature
  space,''
\newblock in {\em {IEEE} Conference on Computer Vision and Pattern Recognition,
  {CVPR} 2021}.

\bibitem{HEVC_CTC}
Frank Bossen,
\newblock ``Common test conditions and software reference configurations,''
\newblock in {\em JCTVC-L1100}, 2013.

\bibitem{9287049}
Théo Ladune, Pierrick Philippe, Wassim Hamidouche, Lu~Zhang, and Olivier
  Déforges,
\newblock ``Optical flow and mode selection for learning-based video coding,''
\newblock in {\em 2020 IEEE 22nd International Workshop on Multimedia Signal
  Processing (MMSP)}.

\bibitem{CLIC21}
``Workshop and challenge on learned image compression,
  https://www.compression.cc/,'' .

\bibitem{AIVCwebsite}
Th{{\'{e}}}o Ladune and Pierrick Philippe,
\newblock ``Artificial intelligence based video coding,'' 2021,
\newblock \url{https://orange-opensource.github.io/AIVC/}.

\bibitem{thesistheo}
Th{{\'{e}}}o Ladune,
\newblock {\em Design of Learned Video Coding Schemes},
\newblock Phd thesis, INSA Rennes, 2021.

\bibitem{balle2018variational}
Johannes Ball{\'{e}}, David Minnen, Saurabh Singh, Sung~Jin Hwang, and Nick
  Johnston,
\newblock ``Variational image compression with a scale hyperprior,''
\newblock in {\em 6th International Conference on Learning Representations,
  {ICLR} 2018}.

\bibitem{Cui_2021_CVPR}
Ze~Cui, Jing Wang, Shangyin Gao, Tiansheng Guo, Yihui Feng, and Bo~Bai,
\newblock ``Asymmetric gained deep image compression with continuous rate
  adaptation,''
\newblock in {\em Proceedings of the IEEE/CVF Conference on Computer Vision and
  Pattern Recognition (CVPR)}, 2021.

\bibitem{hosu2017konstanz}
Vlad Hosu, Franz Hahn, Mohsen Jenadeleh, Hanhe Lin, Hui Men, Tam{\'a}s
  Szir{\'a}nyi, Shujun Li, and Dietmar Saupe,
\newblock ``The konstanz natural video database (konvid-1k),''
\newblock in {\em 2017 Ninth International Conference on Quality of Multimedia
  Experience (QoMEX)}.

\bibitem{DBLP:conf/iclr/YangYMM21}
Ruihan Yang, Yibo Yang, Joseph Marino, and Stephan Mandt,
\newblock ``Hierarchical autoregressive modeling for neural video
  compression,''
\newblock in {\em 9th International Conference on Learning Representations,
  {ICLR} 2021}.

\bibitem{DBLP:conf/mmsys/MercatVV20}
Alexandre Mercat, Marko Viitanen, and Jarno Vanne,
\newblock ``{UVG} dataset: 50/120fps 4k sequences for video codec analysis and
  development,''
\newblock in {\em Proceedings of the 11th {ACM} Multimedia Systems Conference,
  MMSys 2020}.

\bibitem{DBLP:conf/accv/GolinskiPYSC20}
Adam Golinski, Reza Pourreza, Yang Yang, Guillaume Sauti{\`{e}}re, and Taco~S.
  Cohen,
\newblock ``Feedback recurrent autoencoder for video compression,''
\newblock in {\em {ACCV} 2020 - 15th Asian Conference on Computer Vision}.

\bibitem{lu:dvc}
Guo Lu, Wanli Ouyang, Dong Xu, Xiaoyun Zhang, Chunlei Cai, and Zhiyong Gao,
\newblock ``Dvc: An end-to-end deep video compression framework,''
\newblock in {\em 2019 IEEE/CVF Conference on Computer Vision and Pattern
  Recognition (CVPR)}.

\end{thebibliography}

\end{document}